\documentclass[sn-nature]{sn-jnl} 

\usepackage{graphicx}%
\usepackage{multirow}%
\usepackage{amsmath,amssymb,amsfonts}%
\usepackage{amsthm}%
\usepackage{mathrsfs}%
\usepackage[title]{appendix}%
\usepackage{xcolor}%
\usepackage{textcomp}%
\usepackage{manyfoot}%
\usepackage{booktabs}%
\usepackage{algorithm}%
\usepackage{algorithmicx}%
\usepackage{algpseudocode}%
\usepackage{listings}%
\usepackage{appendix}
\usepackage{tcolorbox}

\theoremstyle{thmstyleone}

\theoremstyle{thmstyletwo}%

\theoremstyle{thmstylethree}%

\begin{document}

\title[Article Title]{
Strategic Multi-Agent Learning for Interpretable Action Valuation of All Players in Football
}

\author[1]{\fnm{Kenjiro} \sur{Ide}}\email{ide.kenjiro@g.sp.m.is.nagoya-u.ac.jp}
\author[2]{\fnm{Taiga} \sur{Someya}}\email{tsomeya@g.ecc.u-tokyo.ac.jp}
\author[3]{\fnm{Kohei} \sur{Kawaguchi}}\email{kkawaguchi@ust.hk}
\author*[1,4]{\fnm{Keisuke} \sur{Fujii}}\email{fujii@i.nagoya-u.ac.jp}

\affil[1]{\orgdiv{Graduate School of Informatics}, \orgname{Nagoya University}, \orgaddress{\city{Nagoya}, \country{Japan}}}

\affil[2]{\orgdiv{Graduate School of Arts and Sciences}, \orgname{The University of Tokyo}, \orgaddress{\city{Tokyo}, \country{Japan}}}

\affil[3]{\orgdiv{Department of Economics}, \orgname{The Hong Kong University of Science and Technology}, \orgaddress{\city{Hong Kong}, \country{China}}}

\affil[4]{\orgdiv{Center for Advanced Intelligence Project}, \orgname{RIKEN}, \orgaddress{\city{Osaka}, \country{Japan}}}

\abstract{
Valuing player actions in football, a complex multi-agent system, requires accounting for strategic interactions among 22 players, including off-ball movements and defensive positioning. Existing reinforcement-learning-based valuation methods commonly incorporate other players through the observed state while retaining independent-agent value formulations. To our knowledge, we introduce the first Markov perfect equilibrium (MPE)-inspired framework for valuing the actions of all 22 players using real football tracking and event data, formulating each possession as a finite-horizon multi-agent dynamic game. To improve interpretability, the framework uses Expandable Decision-Making States (EDMS) and decomposes each Q-value into a successor-feature basis and a linear reward-weight vector. The value basis is estimated through a two-stage procedure combining linear temporal-difference initialization with nonlinear iterative refinement from observed state-action transitions. Using tracking and event data from 95 J1 League matches, we compare the proposed formulation with an independent reinforcement learning baseline under both Position and Velocity States (PVS) and EDMS representations. With EDMS held fixed, the independent baseline assigns the highest value to forward movement in 99.21\% of evaluated off-ball states, whereas the most frequent direction under the proposed formulation accounts for 17.63\%. Qualitative analyses illustrate context-dependent off-ball valuations and highlight limitations of the current state representation for defensive positioning. Overall, the framework provides an MPE-inspired approach to interpretable action valuation in football, enabling analysis under alternative tactical reward weights.
}


\maketitle

\section{Introduction}
\label{sec:introduction}
In invasion-type team sports such as football, player actions cannot be evaluated in isolation because 22 players continuously interact on a shared two-dimensional field.
Early rule-based approaches represented player interactions using geometric or kinematic models of ball access and spatial control \cite{Link2016, taki1996development}. Although these models are intuitive and tactically meaningful, they are typically designed for specific situations or concepts, making it difficult to extend them to a unified valuation of on-ball, off-ball, and defensive actions.

Recent machine-learning approaches have modeled football dynamics by predicting discrete events \cite{decroos2019actions, simpson2022seq2event} or forecasting future player trajectories \cite{teranishi2022evaluation}. These approaches are effective for predicting what is likely to happen next from an observed game state. However, action valuation requires a different perspective: estimating the long-term value associated with different player actions in the current tactical context.

Reinforcement learning provides a natural framework for this problem because action-value functions represent the expected cumulative consequences of decisions over time. Accordingly, deep reinforcement learning (DRL) has been applied to estimate player action values from event and tracking data \cite{nakahara2023action}, enabling on-ball and off-ball actions to be evaluated over temporally extended possession sequences. To improve the interpretability of such models, prior work introduced Expandable Decision-Making States (EDMS) \cite{ide2026expandable}, which represents football-specific tactical concepts such as defensive pressure, passing options, and spatial control. However, existing RL-based football valuation methods \cite{rahimian2022beyond,nakahara2023action,ide2024interpretable, ide2026expandable} typically estimate player values under independent-agent formulations. Although other players are represented in the observed state, the strategic interdependence among their decisions is not explicitly incorporated into the value formulation. This limitation is particularly relevant for off-ball and defensive actions, whose values depend strongly on the positioning and responses of other players.

To address this gap, we formulate football action valuation from a multi-agent dynamic-game perspective \cite{adusumilli2026temporal}, treating each possession as a finite-horizon game among interacting players. The formulation is motivated by Markov Perfect Equilibrium (MPE), a solution concept for dynamic games in which each agent follows a state-dependent strategy that is a best response to the strategies of the other agents. Within this formulation, on-ball attackers, off-ball attackers, and defenders are represented as autonomous agents whose action values are conditioned on the current multi-player game state. Player-specific action values are estimated from observed multi-player state-action transitions, enabling the framework to operate directly on real tracking and event data.

Building on EDMS as the state representation, we introduce interpretability at the value level by decomposing each action value into tactically meaningful components. Specifically, each Q-value is decomposed into a successor-feature basis and a linear reward-weight vector, separating the expected future accumulation of tactical reward features from the weights that determine their relative importance. The successor-feature basis is estimated using a two-stage procedure in which linear TD estimates provide supervision for pretraining a nonlinear value approximator, followed by iterative TD refinement.

The contributions of this study are summarized as follows:
\begin{enumerate}
    \item We formulate football action valuation as a finite-horizon multi-agent dynamic game and introduce an MPE-inspired value formulation for on-ball, off-ball, and defensive actions using observed multi-player transitions.

    \item We develop an interpretable successor-feature representation that separates the expected accumulation of tactical reward components from their predefined weights, enabling component-wise interpretation and post-hoc analysis under alternative tactical preferences. The value basis is estimated through a two-stage procedure combining linear TD estimation with nonlinear iterative refinement.

    \item Using tracking and event data from 95 J1 League matches, we show that the proposed formulation produces substantially less concentrated off-ball action rankings than the independent RL baseline under the same EDMS representation, and qualitative analyses further illustrate context-dependent valuations of off-ball movement and a preliminary analysis of defensive positioning.
\end{enumerate}

\section{Related work}
\label{sec:related_work}

\subsection{Player performance evaluation in football}
\label{ssec:player_performance_evaluation_in_football}
Football action valuation aims to quantify the contribution of player actions to attacking or defensive outcomes. Early approaches focused primarily on shots, most notably through Expected Goals (xG) models, which estimate the probability of scoring from shot location and contextual information \cite{robberechts2020data}. Because xG assigns value only at the point of a shot, subsequent research has sought to evaluate a broader range of actions occurring before a scoring opportunity.

A major extension has been the valuation of on-ball actions and ball progression throughout a possession.
VAEP (Valuing Actions by Estimating Probabilities) \cite{decroos2019actions}, for example, assigns values to actions such as passes and dribbles according to their effects on the probabilities of scoring and conceding.
Expected Threat (xT) \cite{singh2018introducing, rudd2011framework} similarly assigns value to pitch locations and evaluates actions according to how they move the ball into more threatening areas.
Related approaches have also evaluated passes using quantities such as interception probability and expected attacking progression \cite{power2017not}.
Together, these methods extend valuation beyond shots by quantifying how on-ball actions change the attacking or defensive value of a possession.

The availability of tracking data further extended football valuation beyond ball-centric events by providing the positions and movements of all players.
The Expected Possession Value (EPV) model \cite{fernandez2019decomposing} estimates possession value from spatiotemporal player configurations, enabling contextual evaluation that incorporates off-ball positioning and player interactions.
This shift toward tracking-based modeling makes it possible to evaluate spatial contributions that are difficult to capture from event data alone.

Tracking data have also enabled explicit quantification of spatial and defensive contributions that occur away from the ball. Spatial-control approaches range from Voronoi-based representations \cite{taki1996development, fujimura2005geometric} to dynamic models incorporating player velocity and acceleration \cite{brefeld2019probabilistic}, while other methods quantify space creation for teammates \cite{fernandez2018wide}. Defensive valuation has similarly been studied through models such as GVDEP \cite{umemoto2022location}, which evaluates how defensive configurations affect the risk of conceding, and models of ball recovery and opponent attacking opportunities \cite{toda2022evaluation}.

Recent deep-learning studies have instead focused on predicting future play, including the next event \cite{simpson2022seq2event}, its time and location \cite{yeung2023transformer}, and future player trajectories \cite{le2017coordinated, fujii2024decentralized}. Such predictions have also been incorporated into play-evaluation frameworks \cite{teranishi2022evaluation, fujii2024estimating}, while systems such as TacticAI \cite{wang2024tacticai} use learned models to support tactical recommendations in specific situations. These approaches demonstrate the value of modeling future game dynamics, but their primary objectives differ from the player-level action-value estimation considered in this study.

\subsection{Reinforcement learning in sports}
\label{ssec:reinforcement_learning_in_sports}
Reinforcement learning (RL) has been applied to sports both in simulation environments for learning decision policies and in data-driven settings for evaluating or inferring player decisions.
In simulation-based research, football has become a prominent testbed for RL through environments such as Google Research Football (GRF) \cite{kurach2020google}. GRF provides a standardized environment for training and evaluating agents and has supported the development of methods for learning cooperative and competitive behaviors. However, policies learned in simulation do not necessarily reflect the decision patterns observed in real professional matches \cite{scott2022how,fujii2024adaptive,fujii2025machine}. Advanced approaches such as curriculum learning and self-play have therefore been explored to learn increasingly sophisticated football strategies \cite{lin2023tizero}.

In parallel, data-driven RL approaches use observations from real matches to estimate action values, evaluate decisions, or infer latent strategies. Early work in ice hockey used play-by-play data to construct Markov-game-based models for context-dependent player action valuation \cite{Liu2018DeepRL}. RL-based player evaluation has also been explored in individual and racket sports such as badminton \cite{ding2022deep}. More recent studies have used real match data to evaluate alternative decisions or infer decision-making strategies. For example, Rahimian et al. \cite{rahimian2022beyond} evaluate alternative player actions in terms of expected possession value, whereas inverse reinforcement learning has been used to infer latent reward structures or strategies from observed offensive and defensive behavior \cite{rahimian2022inferring,luo2020inverse}.

Most directly related to our study, Nakahara et al. \cite{nakahara2023action} model individual football players within a multi-agent reinforcement learning framework using real tracking data, enabling action valuation for both on-ball and off-ball players. Although their framework includes multi-player information in the observed state, player-specific values are estimated under an independent-agent formulation, without explicitly introducing strategic interdependence into the value equations. Our work instead formulates football action valuation from an MPE-inspired dynamic-game perspective. Within this formulation, we use the previously proposed EDMS \cite{ide2026expandable} as a tactically interpretable state representation and combine it with successor-feature-based value decomposition.

\subsection{Game theory in sports}
\label{ssec:game_theory_in_sports}
Competitive sports naturally involve strategic interactions because the value of one player's action depends on the actions and responses of teammates and opponents. Game-theoretic approaches have therefore been used to study decision-making in sports, ranging from isolated one-shot situations to sequential tactical decisions.

Static game-theoretic studies have primarily focused on well-defined situations such as penalty kicks. Because a penalty kick can be modeled as a strategic interaction between the kicker and goalkeeper, several studies have examined whether professional players' behavior is consistent with mixed-strategy Nash equilibrium \cite{palacios2003professionals,chiappori2002testing}. Other studies have investigated deviations from purely rational behavior by considering psychological and behavioral factors \cite{bar2009penalty}. Cooperative game theory has also been applied to quantify individual contributions to team performance, for example through Shapley-value-based evaluation \cite{auer2015evaluation}.

Beyond one-shot interactions, several studies have modeled the temporal evolution of football using dynamic mathematical or game-theoretic frameworks. Physics-inspired approaches have represented interactions among players through stochastic dynamical systems \cite{chacoma2021stochastic}, while models such as Soccer-Decoder describe match progression using differential equations \cite{ferrarini2014new}. These approaches provide useful descriptions of collective or match-level dynamics, but they generally operate at a coarser level than individual action valuation.

Game-theoretic approaches have also been applied more directly to tactical decision-making under uncertainty.
Beal et al. \cite{Beal2020OptimisingGT} modeled pre-match tactical choices using Bayesian games and in-game tactical adjustments using stochastic games. Lennartsson et al. \cite{lennartsson2015game} formulated defensive decision-making through a scoring-probability potential field, while other studies have modeled opponent strategies and searched for counter-strategies in simulated environments \cite{milani2007game}. These studies demonstrate the usefulness of strategic modeling in sports, but applications to fine-grained player-level decisions in real tracking data remain limited.

A central difficulty is the size and continuity of the football state and action spaces. In real matches, players repeatedly make local decisions based on the positions and movements of teammates and opponents, resulting in a high-dimensional sequential decision problem. Explicitly computing equilibrium strategies at this resolution would generally require strong simplifications or substantial computational resources.

Our study addresses this setting from an MPE-inspired dynamic-game perspective. Rather than attempting to compute an exact equilibrium strategy profile, we use MPE as a theoretical motivation for state-dependent multi-agent value equations and estimate player-specific continuation values from observed tracking and event data. By combining this formulation with EDMS and successor-feature-based value decomposition, the framework provides tactically interpretable action values for on-ball, off-ball, and defensive players.

\section{Formulation}
\label{sec:model}
In this section, we formulate a football possession sequence as a finite-horizon dynamic game among multiple decision-making agents. Section~\ref{ssec:problem_setting} defines the state, action, transition, and reward setting. Section~\ref{ssec:mpe} introduces MPE as a motivating solution concept and clarifies how its associated value equations are approximated without computing an exact equilibrium strategy profile. Section~\ref{ssec:value_function_reward_design} defines the tactical reward features and the successor-feature-based value representation. The corresponding two-stage estimation procedure is presented in Section~\ref{sec:method}.

\subsection{Problem Setting}
\label{ssec:problem_setting}
Each football possession sequence is formulated as a finite-horizon stochastic game involving a set of players $I = \{1, \ldots, N\}$. The corresponding joint feasible action set is $A(s_t)=\prod_{i\in I} A_i(s_t)$. In the empirical implementation, $N=22$, corresponding to 11 players on the team in possession and 11 players on the opposing team. The index $i$ denotes a focal-player slot within a possession rather than a persistent player identity across matches. Each possession contains time steps $t=0,\ldots,T$, where the terminal time $T$ may differ across possessions. At each time step $t$, the game is described by a state $s_t \in S$, which is constructed from tracking and event data. Each player $i$ selects an action $a_{i,t} \in A_i(s_t)$, and the joint action is denoted by $a_t = (a_{1,t}, \ldots, a_{N,t})$. The next state is governed by the transition probability $P(s_{t+1} \mid s_t, a_t)$, which depends on the actions of all players.

We consider two state representations: PVS, which describes the physical configuration of the players and ball, and EDMS, which represents football-specific tactical information.
The definitions of PVS and EDMS follow \cite{nakahara2023action,ide2026expandable}.
Table~\ref{tab:state_representations} summarizes their dimensionalities and main variables.

\begin{table}[htbp]
    \centering
    \caption{Summary of the state representations used in the experiments.}
    \label{tab:state_representations}
    \begin{tabular}{lcp{7cm}}
        \hline
        State representation & Dimension & Main variables \\
        \hline
        PVS & 92 &
        Player and ball positions and velocities. \\
        EDMS & 74 &
        Pressure, space, passing options, shooting opportunities,
        offside-line information, and possession context. \\
        \hline
    \end{tabular}
\end{table}

We use the discrete action representation introduced in previous studies \cite{nakahara2023action, ide2026expandable}, consisting of 15 feasible action classes. As summarized in Table~\ref{tab:action_classes}, the action space comprises five on-ball attacking actions, one aggregated defensive action, and nine off-ball actions. A separate padding label is used during sequence batching and is not included in the feasible action space.

\begin{table}[htbp]
    \centering
    \caption{Summary of the feasible discrete action classes.}
    \label{tab:action_classes}
    \begin{tabular}{lcp{6cm}}
        \hline
        Action category & No. of classes & Actions \\
        \hline
        On-ball attacking actions & 5 &
        Pass, through pass, shot, cross, and dribble \\
        Defensive action & 1 &
        Tackle, interception, clearance, or block, aggregated into one defensive-action class \\
        Off-ball actions & 9 &
        Movement in eight directions and idle \\
        \hline
        Total & 15 & \\
        \hline
    \end{tabular}
\end{table}

The feasible action set $A_i(s_t)$ depends on the player's role at state $s_t$.
As summarized in Table~\ref{tab:action_mask}, the attacking ball carrier can select only on-ball attacking actions, attacking off-ball players can select among the nine off-ball actions, and defending players can select either off-ball movements or the aggregated defensive action.

\begin{table}[htbp]
    \centering
    \caption{State-dependent action masking according to player role.
    Actions not listed as feasible are masked.}
    \vspace{2mm}
    \label{tab:action_mask}
    \begin{tabular}{lp{8cm}}
        \hline
        Player role & Feasible actions \\
        \hline
        Attacking ball carrier &
        Pass, through pass, shot, cross, and dribble \\
        
        Attacking off-ball player &
        Movement in eight directions and idle \\
        
        Defending player &
        Movement in eight directions, idle, and defensive action \\
        \hline
    \end{tabular}
\end{table}

Each player receives an immediate reward $r_i(s_t,a_t)$ determined by the current state and joint action.
In this study, the reward is represented through a linear combination of tactical reward features, as defined in Section~\ref{ssec:value_function_reward_design}.
The state-dependent Markov policies and finite-horizon value functions associated with this stochastic-game formulation are introduced in Section~\ref{ssec:mpe}.

\subsection{Dynamic-Game Formulation Inspired by Markov Perfect Equilibrium}
\label{ssec:mpe}
In this study, MPE serves as the motivating solution concept for modeling state-dependent strategic behavior in a dynamic multi-agent setting.

In dynamic multi-stage games, a Nash equilibrium may rely on continuation strategies that are not credible after histories that are not reached on the equilibrium path. Subgame-perfect equilibrium (SPE) strengthens Nash equilibrium by requiring the strategy profile to constitute a Nash equilibrium in every subgame, thereby ruling out such non-credible threats. However, a general SPE may allow strategies to depend on the full history of play. In environments such as football, where the state space is large and continuous, representing history-dependent strategies is computationally impractical. We therefore adopt an MPE perspective and restrict attention to Markov strategies, in which each player's strategy depends only on the current game state.

For each player $i \in I$ and time step $t$, a Markov strategy $\pi_{i,t}$ assigns a probability distribution over the feasible action set:
\begin{equation}
    \pi_{i,t}(\cdot \mid s_t) \in \Delta(A_i(s_t)).
\end{equation}

An MPE is a Markov strategy profile $\pi^*=\{\pi_{i,t}^*\}_{i\in I,t=0,\ldots,T}$ such that no player can improve its expected return by unilaterally changing its strategy while the strategies of the other players remain fixed. Formally, for every player $i$, state $s_t \in S$, and alternative Markov strategy $\tilde{\pi}_i$, the following condition holds:
\begin{equation}
    V_{i,t}^{\pi^*}(s_t)
    \ge
    V_{i,t}^{(\tilde{\pi}_i,\pi_{-i}^*)}(s_t),
    \qquad
    \forall i,\ \forall t,\ \forall s_t,\ \forall \tilde{\pi}_i.
\end{equation}

The finite-horizon value function for player $i$ under strategy profile $\pi$ is
\begin{equation}
    V_{i,t}^{\pi}(s_t)
    =
    \mathbb{E}_{\pi}
    \left[
        \sum_{k=t}^{T}
        \gamma^{k-t} r_i(s_k,a_k)
        \,\middle|\,
        s_t
    \right].
\end{equation}

In this study, the discount factor is set to $\gamma=1$.
Because each possession has a finite terminal time $T$, the return remains finite without discounting, and rewards occurring later in the possession are propagated without temporal attenuation.

From the perspective of player $i$, fixing the Markov strategies of the other players induces a state-dependent stochastic environment in which their action distributions depend on the current state $s_t$. The resulting decision problem can therefore be interpreted as an induced Markov decision process for player $i$. The MPE condition requires the state-dependent strategies of all players to be mutually consistent best responses.

The empirical implementation does not explicitly estimate the Markov policies $\pi_{i,t}$ or verify an equilibrium strategy profile. Instead, observed player actions are used to construct Bellman-style continuation targets from the state-action transitions in the dataset.
Specifically, continuation values are bootstrapped using the next actions observed in each possession rather than actions obtained from an explicit best-response or greedy-policy update. 
Information about the other players enters through the joint game state and the realized transition in the observed data, without explicit enumeration or optimization over all joint action profiles.

Accordingly, MPE provides the theoretical motivation for the state-dependent multi-agent value formulation, but the estimation procedure should not be interpreted as an algorithm for computing an MPE.
Rather, it estimates player-specific continuation values under observed behavior within the proposed dynamic-game representation.

\subsection{Interpretable Value Representation and Reward Design}
\label{ssec:value_function_reward_design}
To obtain an interpretable action-value representation, we decompose each Q-value into a value basis and a reward-weight vector rather than representing it as a single scalar output.
The value basis represents the expected future accumulation of tactically meaningful reward features, whereas the reward-weight vector specifies their relative contributions to the final action value.

For this purpose, the immediate reward of player $i$ is defined as a linear combination of reward features:
\begin{equation}
r_i(s_t, a_t) = z_i(s_t, a_t)^{\top} \theta,
\end{equation}
where $z_i(s_t,a_t)$ denotes the reward-feature vector and $\theta$ is the reward-weight vector. 
For attacking players, the reward features may depend on the focal player's action, whereas all defending players receive the same team-level defensive reward features at a given time step.

In this study, we introduce $d=3$ features corresponding to shot value, possession termination, and attacking-third ball occupancy.
Let $V(x_t,y_t)$ denote the Expected Possession Value (EPV) \footnote{https://github.com/Friends-of-Tracking-Data-FoTD/LaurieOnTracking} obtained from the EPV grid at the ball location $(x_t,y_t)$. For a shot at time $t$, we set the shot value to $v_t=V(x_t, y_t)$.
Let $S_{i,t}=\mathbf{1}\{a_{i,t}=\mathrm{shot}\}$ and let $E_t$ indicate the final time step of a possession segment.
Let $L$ denote the pitch length, with the origin located at the center of the pitch and the attacking direction aligned with the positive x-axis. In our setting, $L=105$~m, so the attacking-third boundary is $x=L/6=17.5$ m.
Let $\mathcal{I}_{\mathrm{att}}$ denote the set of attacking players.
We define

\begin{equation}
C_t = E_t\mathbf{1} \left\{\sum_{j\in\mathcal{I}_{\mathrm{att}}} S_{j,t}=0\right\}\mathbf{1}\{x_t<L/6\}.
\end{equation}

For an attacking player $i$, the reward-feature vector is defined as

\begin{equation}
z_i(s_t,a_t) = \begin{bmatrix}
v_t S_{i,t} \\
V(-x_t,y_t)C_t \\
\mathbf{1}\{x_t\geq L/6\}
\end{bmatrix}.
\end{equation}

For a defending player $i$, it is defined as

\begin{equation}
z_i(s_t,a_t) = \begin{bmatrix}
-\mathbf{1}\left\{\sum_{j\in\mathcal{I}_{\mathrm{att}}} v_t S_{j,t}>0 \right\} \\
-V(-x_t,y_t)C_t \\
0
\end{bmatrix}.
\end{equation}

Here, $v_t$ denotes the configured shot value.
The three components have the following interpretations:

\begin{itemize}
\item $ z_{i,1} $ (Shot value): Nonzero only at a shot time. The shooting attacker receives $v_t$ (e.g., $V(x_t,y_t)$ under the EPV specification), while all defenders receive $-1$.
\item $ z_{i,2} $ (Possession termination adjustment): Nonzero only at the final time step of a possession that ends without a shot and outside the attacking third. Its magnitude is the opponent-perspective EPV, $V(-x_t,y_t)$, with opposite signs for attackers and defenders.
\item $z_{i,3}$ (Attacking-third ball occupancy): Equals one for every attacking player whenever the ball is located in the attacking third, $x_t\geq L/6$, and zero otherwise. Thus, it represents ball occupancy of the attacking third rather than only the first entry into it.
\end{itemize}

The weight parameters are predefined as $\theta = [1.0, -0.5, 0.2]^\top$.
These weights were set heuristically to reflect the relative importance of the three reward components: shot value, possession termination adjustment, and attacking-third ball occupancy. They were not estimated from data in this study.

Based on the linear reward specification, the action-value function can be decomposed into the inner product of a successor-feature representation and the reward-weight vector.
Under a Markov strategy profile $\pi$, the action-value function for player $i$ is defined as

\begin{equation}
Q_{i,t}^{\pi}(s_t,a_{i,t})
=
h_{i,t}^{\pi}(s_t,a_{i,t})^\top\theta.
\label{eq:q_decomposition}
\end{equation}
where $h_{i,t}^{\pi}(s_t,a_{i,t})$, referred to as the value basis or successor-feature vector, represents the expected discounted accumulation of the reward features obtained after taking action $a_{i,t}$ at state $s_t$ and subsequently following the strategy profile $\pi$:

\begin{equation}
h_{i,t}^{\pi}(s_t,a_{i,t})
=
\mathbb{E}^{\pi}
\left[
\sum_{k=t}^{T}
\gamma^{k-t}
z_i(s_k,a_k)
\mid
s_t,a_{i,t}
\right].
\end{equation}

The expectation is taken over future state transitions and the actions of the other players under the strategy profile $\pi$, including the current actions $a_{-i,t}\sim\pi_{-i,t}(\cdot\mid s_t)$.
This equation defines the theoretical successor-feature representation under a Markov strategy profile.

In the empirical estimator, however, the continuation distribution induced by $\pi$ is not explicitly estimated.
Instead, Bellman-style TD targets are constructed from the realized joint state-action transitions in the observed possession sequences.
In particular, the continuation term uses the next action actually observed in the data rather than an action selected by an explicitly estimated greedy or best-response policy.
Thus, the learned value basis should be interpreted as an estimate of feature continuation values under the observed behavior represented in the dataset.

The linear decomposition separates the expected accumulation of tactical reward features, represented by $h_{i,t}$, from their relative weighting, represented by $\theta$.
This makes the estimated action value interpretable in terms of the contribution of each reward component.
Once the value basis has been estimated, alternative reward weights can be applied post hoc to examine how different tactical preferences alter local action rankings.
Because the learned value basis is held fixed and the TD targets are not recomputed under the alternative reward weights, this reweighting should be interpreted as a sensitivity analysis of the fixed value representation rather than as re-estimation of the value function under a new policy or reward specification.

\section{Method}
\label{sec:method}
\subsection{Learning Objective}
\label{ssec:learning_objective}
Our estimation procedure builds on the temporal-difference framework of Adusumilli and Eckardt \cite{adusumilli2026temporal}, who develop linear semi-gradient and iterative value-function estimators for recursive value terms in dynamic discrete-choice models and dynamic games.
Their primary objective is structural parameter estimation, in which the estimated recursive value terms are subsequently incorporated into a conditional-choice-probability-based pseudo-likelihood.
In contrast, our objective is football action valuation.
We fix the tactical reward-weight vector $\theta$ and directly estimate the successor-feature value basis $h_i$ from observed state-action transitions.

We make two main adaptations to this framework.
First, rather than treating the linear and nonlinear estimators as alternative procedures, we combine them sequentially: linear semi-gradient TD estimates are used as supervision to pretrain a nonlinear value approximator, which is subsequently refined through iterative Bellman-style TD updates.
Second, continuation targets are constructed using the next actions observed in the possession sequences rather than actions selected by a greedy or explicit best-response policy.
Accordingly, the estimation procedure does not compute an equilibrium strategy profile or perform policy optimization; it estimates continuation values under the observed behavior represented in the data.

The resulting learning objective is to estimate a nonlinear successor-feature basis that is consistent with the observed finite-horizon state-action transitions.
We use a two-stage procedure.
In the first stage, a linear semi-gradient TD estimator produces initial successor-feature estimates for the observed state-action pairs.
A randomly initialized multilayer perceptron (MLP) is then pretrained to reproduce these estimates by minimizing squared prediction error; the linear coefficients themselves are not transferred to the network parameters.
In the second stage, the pretrained MLP is iteratively refined using Bellman-style targets constructed from the observed next state-action pairs, allowing nonlinear state-dependent structure in the continuation values to be captured.

\subsection{Linear TD Initialization of Successor Features}
\label{ssec:linear_td_initialization}
In the first stage of learning, the value basis $h_i(s_t,a_{i,t}) \in \mathbb{R}^{d}$ is approximated using a linear model:

\begin{equation}
h_i(s_t,a_{i,t};W_i)
=
W_i^\top \phi_i(s_t,a_{i,t}),
\end{equation}

where $\phi_i(s_t,a_{i,t})\in\mathbb{R}^{p_i}$ denotes the basis-function vector for focal-player slot $i$, and $W_i\in\mathbb{R}^{p_i\times d}$ is the parameter matrix to be learned.
A separate linear basis pipeline is constructed for each of the 22 focal-player slots, corresponding to the 11 attacking and 11 defending slots within each possession rather than to persistent player identities. Let $o_{i,t}$ denote the player-centric state representation for focal player $i$, which contains its relative tactical context, global pitch context, and focal-player location.

For each focal-player slot, the observed action is represented by its integer action ID, and the value basis contains a separate action block for every action category.
Continuous components of $o_{i,t}$ are standardized using the mean and standard deviation estimated from the corresponding training observations. The basis vector is then constructed as

\begin{equation}
\tilde{o}_{i,t} = \operatorname{Standardize}_i(o_{i,t}).
\end{equation}

Let $c_{i,t}$ denote the role category of focal player $i$ at time $t$.
The basis vector is constructed as 
\begin{equation}
\phi_i(s_t,a_{i,t}) = \mathbf{e}_{(c_{i,t},a_{i,t})} \otimes
\begin{bmatrix}
1 \\
\tilde{o}_{i,t}
\end{bmatrix},
\end{equation}

where $\mathbf{e}_{(c_{i,t},a_{i,t})}$ is a one-hot vector identifying
the corresponding role--action combination. This block-structured representation allows the coefficients of the state variables to vary across role-action combinations. It does not include higher-order state-state polynomial terms.

In this study, $d=3$, corresponding to the three reward features defined in Section~\ref{ssec:value_function_reward_design}. Each column of $W_i$ therefore estimates the expected discounted accumulation of one reward feature.

For an observed transition
$(s_t,a_t,z_i(s_t,a_t),s_{t+1},a_{i,t+1})$,
the TD target is
\begin{equation}
Y_{i,t} = z_i(s_t,a_t) + \gamma h_i(s_{t+1},a_{i,t+1};W_i),
\end{equation}
where $a_{i,t+1}$ is the next action of focal player $i$ observed in the data.
At the terminal time step,
\begin{equation}
Y_{i,T} = z_i(s_T,a_T),
\end{equation}
because the continuation value is set to zero.

The corresponding TD residual is
\begin{equation}
\delta_{i,t} = Y_{i,t} - h_i(s_t,a_{i,t};W_i).
\end{equation}

Using a semi-gradient update, the TD target is treated as fixed with respect to $W_i$ during each parameter update:
\begin{equation}
W_i \leftarrow W_i + \alpha\phi_i(s_t,a_{i,t}) \delta_{i,t}^{\top},
\end{equation}
where $\alpha$ denotes the learning rate.

For possessions ending with a shot, the shot-value component ($z_{i,1}$) is thus assigned directly on the terminal row. 
For possessions ending without a shot, the possession-termination component ($z_{i,2}$) is assigned on the terminal row and enters the Bellman target directly. 
The attacking-third ball occupancy component ($z_{i,3}$) is assigned at every time step in which the ball is in the attacking third, including the terminal time step when applicable. 
These immediate reward features are propagated to preceding states through the estimated continuation values.

The resulting linear TD predictions are then used as supervision for pretraining the nonlinear value approximator.

\subsection{Approximate Value Iteration (AVI) for Nonlinear Value-Basis Refinement}
\label{ssec:avi}
In the second stage, we refine the successor-feature value basis using a nonlinear function approximator implemented as a MLP.
The linear TD predictions obtained in the first stage are used as supervision for pretraining the MLP.

Following the approximate value-iteration terminology of the underlying TD framework \cite{adusumilli2026temporal}, we refer to this iterative nonlinear refinement as AVI.
In our implementation, however, the continuation target uses the next action observed in the data rather than an action selected by a greedy or explicit best-response policy.
Accordingly, the procedure performs iterative fitted value-basis refinement under observed continuation behavior rather than policy improvement through a Bellman optimality operator.

The learning procedure consists of the following steps:

\begin{enumerate}

    \item \textbf{Pretraining:}
    The MLP parameters $\psi$ are randomly initialized and pretrained to reproduce the linear TD predictions obtained in the first stage by minimizing the squared prediction error.
    The resulting parameters are used to initialize the subsequent AVI refinement.
    The linear coefficients $W_i$ are therefore not directly transferred to the MLP parameters.

    \item \textbf{AVI iteration $(k=0,1,2,\ldots)$:}
    At iteration $k$, for each observed transition $(s_t,a_t,z_i(s_t,a_t),s_{t+1},a^{\mathrm{obs}}_{i,t+1})$, the target is constructed as

    \begin{equation}
    y^{(k)}_{i,t} = z_i(s_t,a_t) + \gamma h_i(s_{t+1},a^{\mathrm{obs}}_{i,t+1};\psi_k),
    \end{equation}

    where $a^{\mathrm{obs}}_{i,t+1}$ denotes the next action of focal player $i$ observed in the possession sequence.

    For a terminal transition, the continuation value is set to zero, giving

    \begin{equation}
    y^{(k)}_{i,T} = z_i(s_T,a_T).
    \end{equation}

    The target $y^{(k)}_{i,t}$ is held fixed while the parameters are updated by minimizing the squared prediction error:

    \begin{equation}
    \psi_{k+1}
    \leftarrow
    \arg\min_{\psi}
    \mathbb{E}_{\mathcal{D}}
    \left[
    \left\|
    y^{(k)}_{i,t}
    -
    h_i(s_t,a_{i,t};\psi)
    \right\|_2^2
    \right],
    \end{equation}

    where $\mathcal{D}$ denotes the observed transition dataset.
    The fitted parameters from iteration $k$ are used to initialize the next iteration.

    \item \textbf{Stopping:}
    The AVI refinement is repeated until the validation-based stopping criterion is met or the predefined maximum number of iterations is reached.

\end{enumerate}

The resulting nonlinear value basis is subsequently used to compute the action value of each feasible action as

\begin{equation}
Q_i(s,a) = h_i(s,a)^\top\theta.
\end{equation}

Importantly, greedy actions are derived only at the post-training inference stage for action-ranking analysis.
They are not used in the AVI targets and therefore do not influence the iterative value-basis estimation.
Thus, although we retain the term AVI to describe the iterative fitted-value procedure, the empirical implementation estimates continuation values under observed behavior rather than performing greedy policy improvement or explicitly computing an equilibrium strategy profile.

\section{Experimental Results}
\label{sec:quantitative_eval}

\subsection{Experimental Setup}
\label{ssec:experimental_setup}

We evaluated the proposed framework using tracking and event data from 95 Meiji Yasuda J1 League matches from the 2023 and 2024 seasons, provided by Data Stadium Inc. The tracking data contain the positions of all players and the ball at 25 fps, while the event data contain annotations of actions such as passes, shots, tackles, clearances, and interceptions. The tracking data were resampled to 10 fps for model estimation. The dataset was split at the match level into 50 training matches from the 2023 season and 45 matches from the 2024 season of which 5 were used for validation and 40 for testing.

The tracking data were preprocessed by interpolating missing values and computing the velocity and acceleration of each player and the ball. The tracking and event data were then synchronized using timestamps. The attacking direction was normalized so that the attacking team always moves from left to right. Each match was segmented into possession sequences using the possession sequence identifier, and sequences with lengths between 30 and 300 time steps at 10 fps were retained.

After preprocessing, the dataset contained 10,308 possession sequences and 1,561,255 time steps. Independently, the total duration of the extracted possession sequences, calculated from the difference between each sequence's start and end frames, was 154,113.3 s (42.81 h). Specifically, the training, validation, and test sets contained 4,589, 918, and 4,801 possession sequences, respectively, comprising 684,605, 149,629, and 727,021 time steps. Their total durations were 67,565.0 s (18.77 h), 14,783.6 s (4.11 h), and 71,764.7 s (19.93 h), respectively.

To examine effects of both the learning formulation and the state representation, we evaluated both the independent RL baseline and the MPE-inspired formulation using PVS and EDMS, resulting in four configurations: Independent+PVS \cite{nakahara2023action}, Independent+EDMS \cite{ide2026expandable}, MPE-inspired+PVS, and MPE-inspired+EDMS. 

The independent RL baseline estimates scalar Q-values using the previously proposed GRU-based architecture, whereas the proposed formulation estimates a three-dimensional successor-feature basis using the two-stage procedure described in Section \ref{sec:method}. For the proposed model, observations from all 22 focal-player slots are pooled to train a shared nonlinear value approximator.

The independent baseline uses a single-layer GRU sequence encoder with an input size of 64 and a hidden size of 64. The baseline is trained using the Adam optimizer with a learning rate of $10^{-3}$ and weight decay of $10^{-5}$. The discount factor is set to $1.0$, the regularization weight to $10^{-5}$, and the supervised action-loss weight to $0.005$.

For the proposed formulation, the successor-feature basis was approximated using an MLP with hidden-layer sizes of 64 and 32. The linear semi-gradient TD estimator was trained for one epoch with a learning rate of $10^{-3}$, and its predictions were used to pretrain the nonlinear value approximator. Nonlinear refinement was then performed for at most 20 iterations, with the value predictor trained for two epochs per iteration using a batch size of 2048. Parameters were warm-started from the preceding iteration, and the learning rate was multiplied by 0.3 at iterations 5, 10, and 15, with a minimum learning rate of $10^{-5}$. Validation attacker TD MSE was used for model selection and early stopping, with patience-based stopping applied after at least five iterations.

\subsection{Evaluation Metrics}
\label{ssec:evaluation_metrics}

We use three quantitative metrics. 
First, we use mean squared temporal-difference error (TD MSE) to assess consistency with the fitted TD targets within each learning formulation. For the MPE-inspired formulation, the TD residual is three-dimensional, corresponding to the successor-feature components: squared residuals are summed across components and averaged across samples and focal-player slots. Because the independent RL baseline defines TD error in a scalar Q-value space whereas the proposed formulation defines it in a successor-feature vector space, absolute TD MSE values are not directly comparable across the two formulations.

Second, the distribution of the highest-valued off-ball actions was analyzed as a diagnostic for action-ranking collapse. Although a more diverse distribution does not by itself prove tactical optimality, it indicates whether the learned action ranking avoids convergence to a trivial dominant movement. 

Third, as an exploratory team-level diagnostic, we examined the association between each team's Q-value summary and its official 2024 J1 League full-season expected goals (xG). Because the Q-value summaries are computed from the held-out tracking-data sample whereas xG is a full-season club-level aggregate, this analysis is interpreted as an exploratory association with season-level attacking performance rather than as a direct validation of the learned action values.
We report Spearman's rank correlation coefficient $\rho$.

\subsection{Quantitative Results}
\label{ssec:results}

\subsubsection{TD Consistency within Each Learning Formulation}
\label{sssec:td_results}

Table~\ref{tab:quantitative_results} reports the TD MSE for the independent RL and MPE-inspired formulations using PVS and EDMS.

\begin{table}[!htbp]
    \centering
    \caption{TD MSE for each learning formulation and state representation.}
    \label{tab:quantitative_results}
    \begin{tabular}{p{3.2cm}p{3.5cm}p{2.2cm}c}
        \hline
        Learning formulation & TD target space & State representation & TD MSE \\
        \hline
        Independent RL~\cite{nakahara2023action}
        & Scalar Q-value
        & PVS
        & $6.6947 \times 10^{-4}$ \\

        Independent RL~\cite{ide2026expandable}
        & Scalar Q-value
        & EDMS
        & $4.5057 \times 10^{-4}$ \\

        MPE-inspired RL
        & Successor features ($d=3$)
        & PVS
        & $1.1229 \times 10^{-2}$ \\

        MPE-inspired RL
        & Successor features ($d=3$)
        & EDMS
        & $1.0876 \times 10^{-2}$ \\
        \hline
    \end{tabular}
\end{table}

Within the independent RL formulation, EDMS yielded a lower TD MSE than PVS ($4.5057\times10^{-4}$ vs.\ $6.6947\times10^{-4}$). The same pattern was observed for the MPE-inspired formulation, for which EDMS also yielded a lower TD MSE than PVS ($1.0876\times10^{-2}$ vs.\ $1.1229\times10^{-2}$).
These within-formulation comparisons indicate that the EDMS provides better consistency with the corresponding fitted TD targets under both learning procedures.

As described in Section~\ref{ssec:evaluation_metrics}, the absolute TD MSE values are not compared across the two learning formulations because their TD targets are defined in different spaces.

\subsubsection{Off-Ball Action Ranking Analysis}
\label{sssec:offball_distribution}
To examine the behavior of alternative-action rankings while holding the state representation fixed, we compare the Independent RL+EDMS and MPE-inspired RL+EDMS configurations.

\begin{figure}[!htbp]
    \centering
    \includegraphics[width=0.95\linewidth]{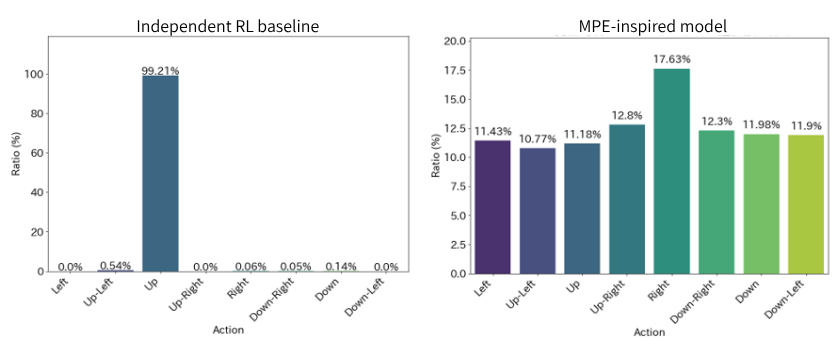}
    \caption{Distribution of highest-valued off-ball actions obtained from post-training Q-value inference. The off-ball action space consists of eight directional movements and an idle action. The idle action was included when determining the highest-valued action but is omitted from the figure for visualization.
    The independent RL baseline assigns the highest Q-value to forward movement in almost all evaluated off-ball states, whereas the proposed MPE-inspired model distributes the highest-valued actions across multiple directions. ``Up'' corresponds to the attacking direction toward the opponent's goal.}
    \label{fig:action_distribution}
\end{figure}

Figure~\ref{fig:action_distribution} shows the distribution of highest-valued off-ball actions. The independent RL baseline assigns the highest Q-value to forward movement in 99.21\% of the evaluated off-ball states, indicating a highly concentrated action ranking. In contrast, the MPE-inspired formulation distributes the highest-valued actions across multiple directions, with the most frequent direction accounting for 17.63\% of the evaluated states.

Because both models use EDMS, this difference is not attributable to the state representation alone. However, the two formulations also differ in model architecture, reward representation, learning objective, and training procedure. Accordingly, the result should be interpreted as a difference between the two learning formulations as a whole rather than as an isolated effect of the MPE-inspired component.

\subsubsection{Exploratory Team-Level Association with Season-Level xG}
\label{sssec:xg_correlation}
We next examine the association between the learned Q-values and season-level attacking performance as an exploratory team-level diagnostic.

\begin{figure}[!htbp]
    \centering
    \includegraphics[width=0.95\linewidth]{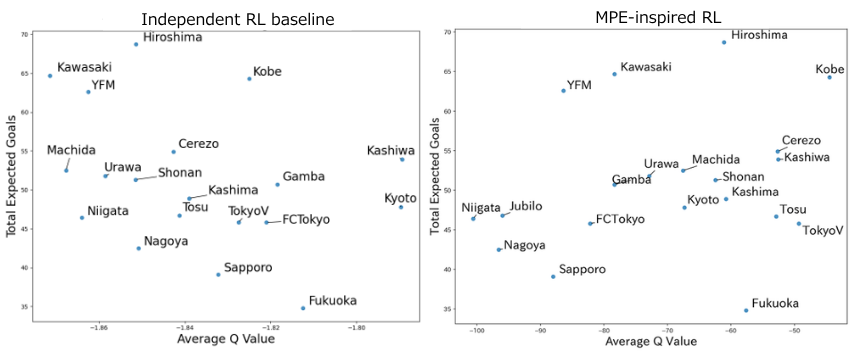}
    \caption{Exploratory association between team-level Q-value summaries computed from the held-out tracking-data sample and official 2024 J1 League full-season xG. Each point represents one of the $n=20$ matched teams. The independent RL baseline shows a negative Spearman rank correlation, whereas the proposed MPE-inspired model shows a weakly positive correlation.}
    \label{fig:xg_qvalue_correlation}
\end{figure}

Figure~\ref{fig:xg_qvalue_correlation} shows the relationship between each team's Q-value summary computed from the 40 held-out test matches and official 2024 J1 League full-season xG.
The xG values were obtained from the official J.League club statistics.\footnote{https://www.jleague.jp/j1/stats/player/2024/expected\_goals/search-list/}
For each team, the Q-value summary was computed by averaging the Q-values of the observed actions over all evaluated player-time-step observations in the held-out test matches involving that team.
Thus, the Q-value summaries and xG values differ in temporal coverage, with the former based on the held-out tracking-data sample and the latter covering the full 2024 season.
Across the $n=20$ matched teams, the independent RL baseline shows a negative correlation ($\rho=-0.37$), whereas the proposed MPE-inspired model shows a weakly positive association ($\rho=0.29$).

These results indicate a difference in the direction of the team-level association between the two learning formulations, rather than demonstrating external validity. Given the small team-level sample and the differing temporal coverage of the two quantities, the results should be interpreted as descriptive. The modest magnitude of the associations may also reflect the fact that xG primarily reflects shot quality, whereas the proposed Q-values represent broader decision-making processes, including build-up play, off-ball movement, and defensive positioning.

\subsection{Qualitative Analysis}
\label{sec:qualitative_analysis}
The quantitative results show that the proposed formulation produces substantially less concentrated off-ball action rankings than the independent RL baseline.
To further examine how these rankings vary across tactical contexts, we analyze representative match situations and visualize the Q-values assigned to feasible player movements.
The selected examples illustrate open-space runs, support movements, scoring opportunities, defensive positioning, and the effect of reward-weight adjustment.

\subsubsection{Evaluation of Off-ball Behavior}
\label{ssec:evaluation_offball_behavior}

We first examine how the proposed model ranks off-ball movements across different tactical contexts. 
The following examples examine whether the model evaluates off-ball actions according to the surrounding tactical context, rather than simply assigning high values to forward movement.

\begin{figure}[!htbp]
    \centering
    \includegraphics[width=0.95\linewidth]{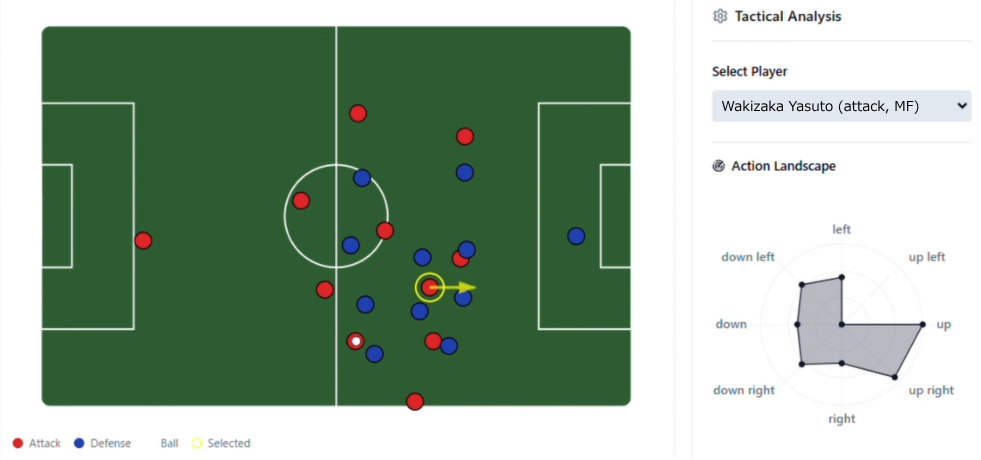}
    \caption{Off-ball Q-value visualization. The left panel shows all players and the ball, with the white marker indicating the ball and the yellow-highlighted player indicating the target player whose Q-values are visualized. The right panel shows the target player's name and a radar chart of Q-values for off-ball movements in eight directions. ``Up'' corresponds to the attacking direction toward the opponent's goal. This example shows off-ball action valuation when open space is available ahead of the target player.}
    \label{fig:front_space}
\end{figure}

Figure~\ref{fig:front_space} shows a situation in which the target off-ball player has open space ahead and the nearby defenders are not positioned to immediately close that space. 
The proposed model assigns relatively high Q-values to forward-oriented movements. 
This valuation is consistent with common tactical principles, because moving into the open space would increase the receiver's separation from defenders and create a more progressive passing option for the ball carrier. 

\begin{figure}[!htbp]
    \centering
    \includegraphics[width=0.95\linewidth]{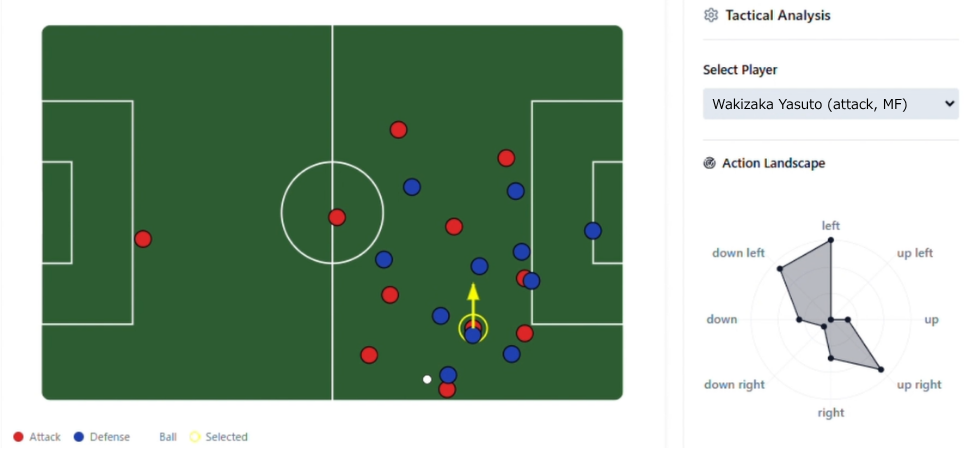}
    \caption{Example of off-ball action valuation in response to a teammate's forward movement. The visualization layout follows Fig.~\ref{fig:front_space}. }
    \label{fig:risk_manage}
\end{figure}

Figure~\ref{fig:risk_manage} shows a situation in which a nearby teammate has already moved into the forward space. 
Unlike the previous example, the target player's highest-valued movements shift toward a supporting position rather than direct forward movement. 
This pattern is consistent with sensitivity to the spatial occupation of nearby teammates, as movements into the already occupied attacking lane receive lower values.
The highest-valued support movement could provide a safer passing option for the ball carrier and maintain the continuity of the possession.

\begin{figure}[!htbp]
    \centering
    \includegraphics[width=0.95\linewidth]{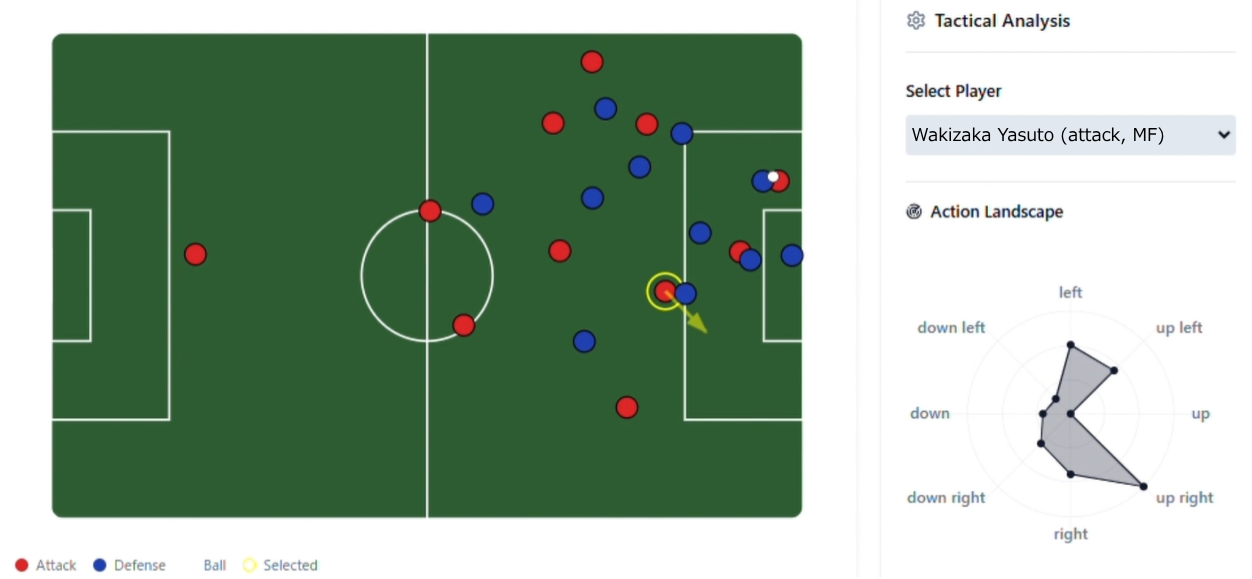}
    \caption{Example of off-ball action valuation during a scoring opportunity. The visualization layout follows Fig.~\ref{fig:front_space}. }
    \label{fig:predict_goal}
\end{figure}

Figure~\ref{fig:predict_goal} presents a situation near the opponent's goal, where small positional changes can directly affect the quality of the attacking opportunity. 
The proposed model assigns high Q-values to movements that improve the target player's relationship with the ball, the goal, and surrounding defenders. 
In particular, the model favors movements that keep the player available as a passing option while avoiding immediate defensive pressure. 
This suggests that the learned Q-values reflect local attacking geometry, rather than only global indicators such as distance to goal.

These examples illustrate that the proposed model can assign Q-values that reflect contextual off-ball decision-making.
Unlike the independent reinforcement learning baseline, whose highest-valued off-ball actions are strongly concentrated on forward movement, the proposed model evaluates multiple movement directions according to the surrounding tactical situation.
These case studies support the interpretation that the proposed learning formulation produces more context-sensitive off-ball action rankings rather than consistently assigning the highest value to forward movement.

\subsubsection{Preliminary Analysis of Defensive Positioning}
\label{ssec:defensive_positioning}

The proposed formulation can also assign action values to defending players.
However, the current state representation was not specifically designed to characterize defensive organization or defensive decision-making.
Although EDMS contains general spatial and interaction-related information, it does not include sufficiently detailed defensive-specific variables for reliable evaluation of defensive positioning.

\begin{figure}[!htbp]
    \centering
    \includegraphics[width=0.95\linewidth]{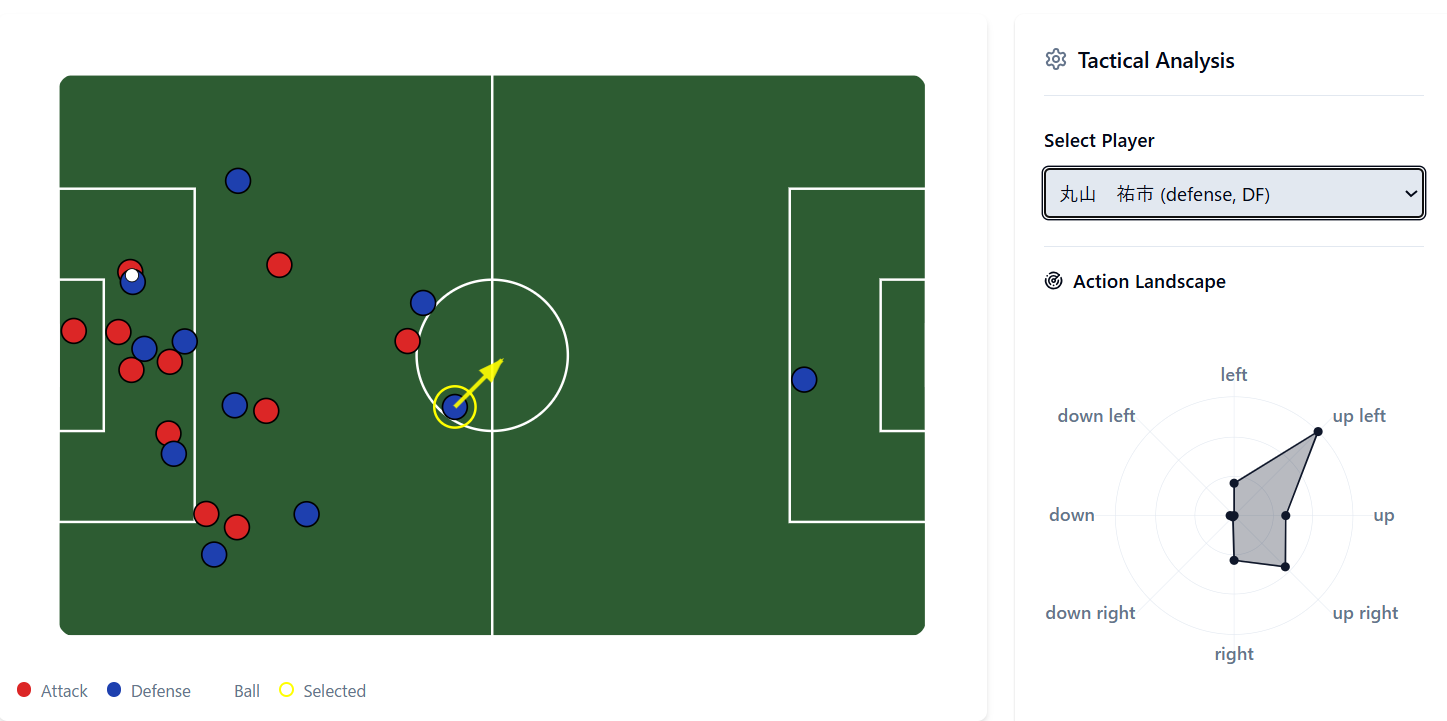}
    \caption{Example of Q-value visualization for a defending player. The visualization layout follows Fig.~\ref{fig:front_space}.}
    \label{fig:defensive_positioning}
\end{figure}

Figure~\ref{fig:defensive_positioning} illustrates the Q-values assigned to alternative movements of a defending player.
While the model produces different values across movement directions, these rankings should not be interpreted as evidence that the current model accurately evaluates defensive positioning.
The present EDMS representation does not explicitly encode defensive concepts such as defensive-line structure, coverage, compactness, or the control of passing lanes from the defender's perspective.

Extending EDMS with defensive-specific state variables could enable more meaningful evaluation of defensive actions and positioning. Such variables could describe, for example, defensive-line organization, spatial coverage, pressure on attackers, and the availability of opponent passing lanes. Developing and validating such defensive representations remains an important direction for future work.

\subsubsection{Reflecting Tactical Preferences through Reward Weight Adjustment}
\label{ssec:reward_weight_adjustment}

We next examine how the learned value representation responds to alternative reward-weight settings. Because the Q-value is computed as $Q_i(s,a)=h_i(s,a)^\top\theta$, the reward weights can be changed after training while keeping the learned successor-feature basis fixed. This should be interpreted as a post-hoc sensitivity analysis rather than as re-estimation of the value model under a new reward function.

\begin{figure}[!htbp]
    \centering
    \includegraphics[width=0.9\linewidth]{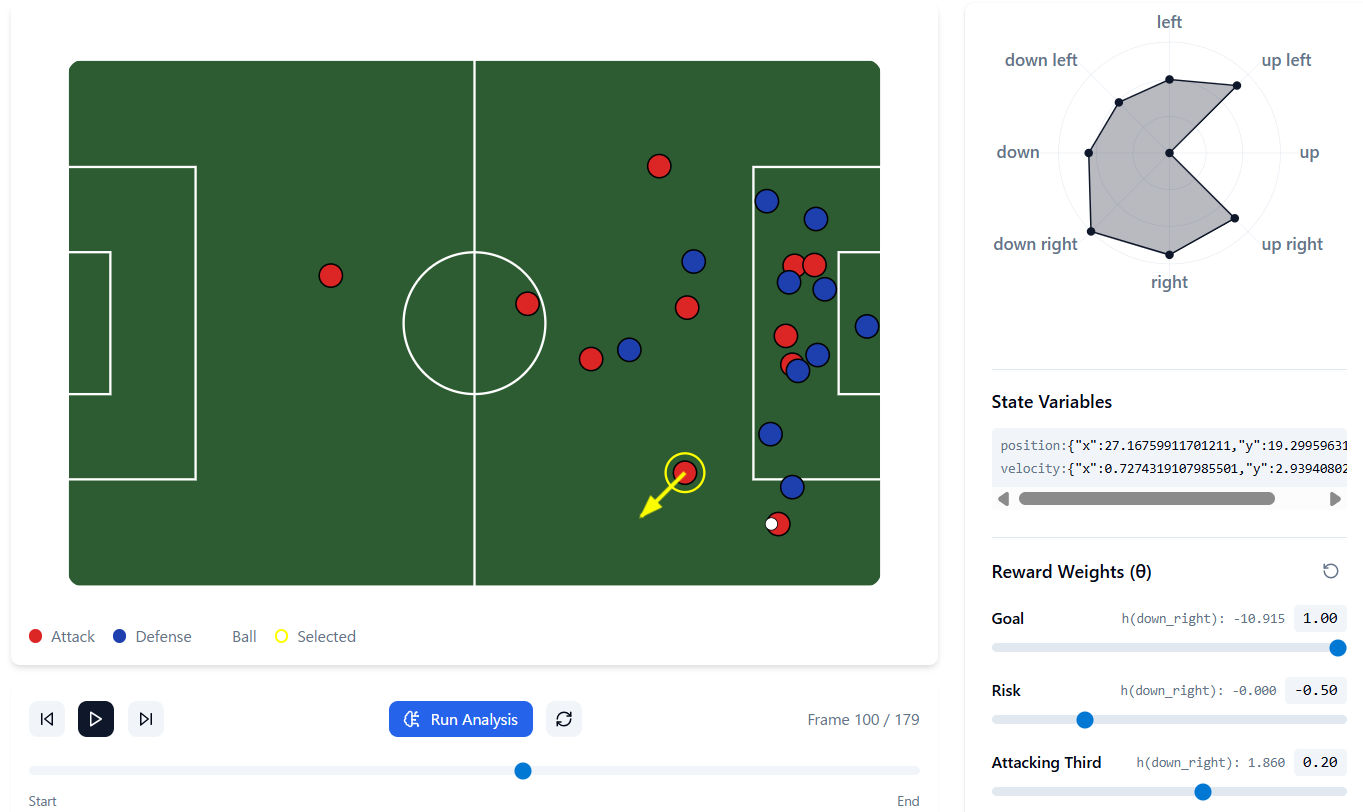}
    \caption{Effect of reward weight adjustment on the highest-valued action. The visualization format follows Fig.~\ref{fig:front_space}. The lower-right panel shows the reward-feature weights $\theta$. Under the default reward weights, $\theta=(1.0,-0.5,0.2)^\top$, the highest-valued action is backward-right.}
    \label{fig:theta_before}
\end{figure}

\begin{figure}[!htbp]
    \centering
    \includegraphics[width=0.9\linewidth]{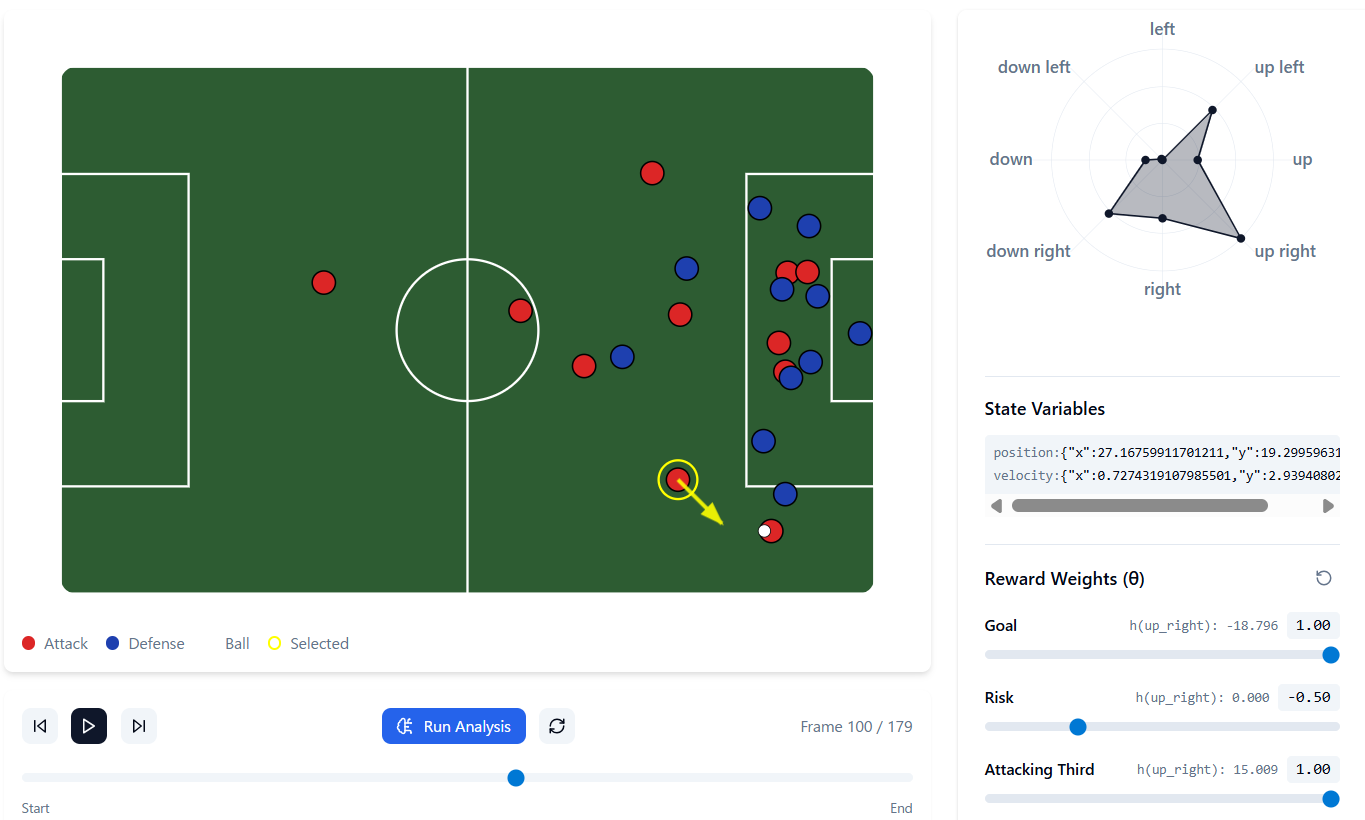}
    \caption{Effect of increasing the attacking-third weight. The visualization format is the same as in Fig.~\ref{fig:theta_before}. After changing the reward weights to $\theta=(1.0,-0.5,1.0)^\top$, the highest-valued action changes from backward-right to forward-right.}
    \label{fig:theta_after}
\end{figure}

Figures~\ref{fig:theta_before} and~\ref{fig:theta_after} compare the highest-valued actions under two different reward-weight settings. 
Figure~\ref{fig:theta_before} shows the result under the default setting, $\theta=(1.0,-0.5,0.2)^\top$. 
Under this setting, the highest-valued action is backward-right.
In contrast, Figure~\ref{fig:theta_after} shows the result after increasing the weight on attacking-third ball occupancy from $0.2$ to $1.0$. 
With this adjustment, the highest-valued action changes toward a more forward-oriented movement.

This comparison illustrates how alternative tactical preferences can change local action rankings through the reward-weight vector.
Because the learned value basis remains fixed, the analysis provides a transparent post-hoc view of how changes in reward priorities affect action valuation without retraining the successor-feature model.

\section{Conclusion}
\label{sec:conclusion}
This study proposed an MPE-inspired framework for interpretable action valuation in football, covering on-ball, off-ball, and defensive players within a finite-horizon multi-agent dynamic-game formulation.
The framework combines a tactically interpretable EDMS state representation with successor-feature-based value decomposition and a two-stage learning procedure consisting of linear TD initialization and nonlinear iterative refinement.
This design enables action values to be decomposed into tactically meaningful reward components and allows post-hoc inspection of how alternative reward preferences affect local action rankings.

Using tracking and event data from 95 J1 League matches, we compared the proposed formulation with an independent RL baseline under both PVS and EDMS state representations.
Within each learning formulation, EDMS yielded lower TD MSE than PVS.
In the EDMS-based comparison, the independent baseline assigned the highest value to forward movement in 99.21\% of the evaluated off-ball states, whereas the most frequent direction under the proposed formulation accounted for 17.63\%.
The team-level association between average Q-values and season-level xG was negative for the independent baseline ($\rho=-0.37$) and weakly positive for the proposed formulation ($\rho=0.29$), although this analysis was exploratory.
Qualitative examples further showed that the proposed formulation produces different off-ball action rankings across tactical contexts and that reward-weight adjustment changes local action rankings without retraining the learned successor-feature basis.

Several aspects of the current study should be considered when interpreting the results.
The present experiments compare the proposed formulation and the independent RL baseline as complete learning systems. Because the two formulations differ in model architecture, reward representation, learning objective, and training procedure, the observed differences should be interpreted at the formulation level rather than as the isolated effect of any single component. Controlled ablation studies using a common backbone and training setup would provide a more fine-grained assessment of the contribution of individual components.
The proposed framework uses MPE as a theoretical basis for state-dependent multi-agent value estimation rather than as an explicit equilibrium-solving procedure.
Continuation values are estimated from observed state-action transitions, and extending the framework toward explicit policy estimation and equilibrium-consistency analysis remains an important direction for future work.
The current EDMS representation was not specifically designed to characterize defensive organization, which limits how confidently defensive-positioning values can be interpreted.
Incorporating defensive-specific variables such as defensive-line structure, spatial coverage, pressure, and passing-lane control will be important for extending the framework to defensive analysis.
The reward weights were manually specified, and their influence should be examined more systematically.
Finally, the current evaluation is based on J1 League data and combines within-formulation TD consistency, action-ranking diagnostics, qualitative case studies, and an exploratory team-level xG analysis.
Validation across additional competitions and outcome-based metrics will be important for assessing generalizability.

Future work should therefore focus on controlled component-wise evaluation, richer state representations for defensive and perceptual context, data-driven or sensitivity-based reward specification, and validation across multiple competitions and outcome-based metrics.
These extensions would help clarify which components of the framework contribute to improved action valuation and support more reliable tactical analysis from real tracking and event data.

\noindent{\textbf{Acknowledgements.} 
The J.League dataset used in this study was provided by the Research Organization of Information and Systems, The Institute of Statistical Mathematics and Data Stadium Inc.
We are grateful to the following supporters of our crowdfunding project (names given in no particular order): 
Eiji Konaka,
Junya Yatabe,
Yu Ukai,
Shotaro Ishihara,
Narita Youkan,
Takashi Hamaji,
Atom Scott,
Daichi Ishii,
Ikuma Uchida,
Yasuyuki Sawada,
Yosuke Yasuda,
Yumu Tabuchi,
Hiroyuki Shindo,
Kazuyuki Saka,
Red Dot Drone Japan,
Yuichiro Someya,
Keita Inoue,
Masato Max Yamamoto,
Rikuhei Umemoto,
Susumu Fukutome,
Takao Toshishige,
Makoto Isono,
Sho Yokoyama,
Leo Nakamura,
Masayuki Fukui, and
Hayato Ohtani.
The author K.I. would like to take this opportunity to thank the ``THERS Make New Standards Program for the Next Generation Researchers.''}

\section*{Declarations}
\begin{itemize}

\item Funding: This work was financially supported by JST SPRING, Grant Number JPMJSP2125. This work was supported by JSPS KAKENHI 23H03282.
\item Competing interests: The authors have no competing interests to declare that are relevant to the content of this article.
\item Ethics approval: Not applicable
\item Consent to participate: Not applicable
\item Consent for publication: Not applicable
\item Availability of data and materials: The event and tracking data used in this study are subject to the terms and licenses of the original data providers and cannot be redistributed by the authors. The released code provides the implementation and experiment scripts, but users must obtain the underlying data from the corresponding data providers under their own licenses.

\item Code availability: The source code for the model implementation, training scripts, experiment scripts, and documentation needed to run the experiments will be made publicly available upon acceptance.

\item Authors' contributions: K.I., T.S., K.K., and K.F. contributed to the study conception and design. Data preparation, modeling, and analysis were performed by K.I. The first draft of the manuscript was written by K.I., and all authors commented on previous versions of the manuscript. All authors read and approved the final manuscript.

\end{itemize}

\bibliography{sn-bibliography}

\newpage
\appendix
\section*{Supplementary Materials}
\section{State Representation}
\label{sec:state_representation}

This appendix provides the detailed definitions of the state representations used in this study. 
We compare two types of state representations: Position and Velocity State (PVS) and Expandable Decision-Making State (EDMS). 
PVS is used as a baseline representation that directly encodes the physical configuration of players and the ball. 
In contrast, EDMS is designed to incorporate tactically interpretable features, such as spatial value, defensive pressure, passing options, and shooting opportunities.

\subsection{PVS: Position and Velocity State}
\label{sec:pvs}

The Position and Velocity State (PVS) is used as the baseline state representation. 
Following previous reinforcement learning-based approaches for football action valuation, PVS represents the game situation using only the physical information of all players and the ball. 
Specifically, for each time step $t$, the state consists of the two-dimensional position and velocity of the ball and all players on the pitch.

Let $p_{i,t} = (x_{i,t}, y_{i,t})$ and $v_{i,t} = (\dot{x}_{i,t}, \dot{y}_{i,t})$ denote the position and velocity of player $i$ at time $t$, respectively. 
Similarly, let $p_{b,t} = (x_{b,t}, y_{b,t})$ and $v_{b,t} = (\dot{x}_{b,t}, \dot{y}_{b,t})$ denote the position and velocity of the ball. 
The PVS representation is then defined as
\begin{equation}
    s^{\mathrm{PVS}}_t =
    \left[
    p_{b,t}, v_{b,t},
    p_{1,t}, v_{1,t},
    \ldots,
    p_{N,t}, v_{N,t}
    \right],
\end{equation}
where $N$ is the number of players considered in the model.

In this study, the attacking direction is normalized so that the attacking team always moves from left to right. 
This normalization allows the model to learn action values under a consistent spatial coordinate system across different possessions and matches.

Although PVS provides a compact and direct description of the physical configuration of the game, it does not explicitly encode tactical relationships among players. 
For example, quantities such as available space, defensive pressure, passing lanes, and shooting angles must be inferred indirectly from raw coordinates and velocities. 
Therefore, while PVS is useful as a baseline representation, it may be limited in terms of sample efficiency and interpretability when modeling complex tactical decision-making in football.

\subsection{EDMS: Expandable Decision-Making State}
\label{sec:edms}

The Expandable Decision-Making State (EDMS) is the proposed state representation used in this study. 
EDMS is designed to represent football-specific tactical information in an interpretable and modular form. 
Unlike PVS, which directly uses raw position and velocity information, EDMS transforms the physical configuration of the pitch into a set of semantically meaningful variables related to space, pressure, passing opportunities, shooting opportunities, and possession context.

The term ``expandable'' indicates that the state representation is not restricted to a fixed set of variables. 
Instead, EDMS is organized as a modular structure, allowing additional tactical or provider-specific variables to be incorporated without changing the overall reinforcement learning framework. 
This property is important for practical football analytics, where the available data and the tactical concepts of interest may differ depending on the league, data provider, or analysis objective.

EDMS consists of two major components: Absolute State and Relative State. 
The Absolute State contains global information that is not defined with respect to a specific target player. 
The Relative State contains player-specific information that changes depending on the evaluated player and the possession context. 
The Relative State is further divided into On-ball State and Off-ball State, depending on whether the target player is in possession of the ball.

\subsubsection{Absolute State}

The Absolute State describes global contextual information of the current game situation. 
In this study, it includes the distance from the ball to each team's offside line and the team formation.

The offside line is defined according to the attacking direction and the positions of the defending players. 
When all attacking players are located in their own half, the halfway line is used as the offside line. 
Otherwise, the offside line is defined by the second-deepest defending player relative to the attacking direction. 
The distance from the ball to this line represents how much vertical space is available behind the defensive line.

The formation variable represents the nominal team formation provided in the data. 
This variable gives coarse tactical context regarding the team structure. 
In the current implementation, formation changes caused by substitutions or in-game tactical adjustments are not explicitly modeled.

\subsubsection{Off-ball State}

The Off-ball State represents the tactical situation of an attacking player who does not possess the ball. 
For each off-ball player, EDMS computes variables that describe passing accessibility, defensive pressure, and spatial value.

First, the distance between the ball and the off-ball player is computed as
\begin{equation}
    d^{\mathrm{ball}}_{i,t}
    =
    \|p_{i,t} - p_{b,t}\|_2.
\end{equation}
This variable captures how easily the player can be involved in the current possession.

Second, EDMS computes the time required for the nearest opposing player to reach the off-ball player. 
This variable is denoted as
\begin{equation}
    \tau^{\mathrm{player}}_{i,t}.
\end{equation}
A larger value indicates that the player is under weaker defensive pressure and has more time to receive or control the ball.

Third, EDMS computes the time required for the nearest opposing player to reach the passing lane from the ball to the off-ball player. 
This variable is denoted as
\begin{equation}
    \tau^{\mathrm{passline}}_{i,t}.
\end{equation}
A larger value indicates that the passing lane is less likely to be intercepted.

Fourth, EDMS includes a space score, which quantifies the value of the space occupied or controlled by the player. 
The space score is computed by combining the player's space control with the tactical importance of the corresponding pitch area. 
The spatial importance is designed to be higher near the opponent's goal and in central areas of the pitch. 
Thus, the space score assigns high values to regions that are both controllable and tactically dangerous.

\begin{figure}[t]
    \centering
    \includegraphics[width=0.95\linewidth]{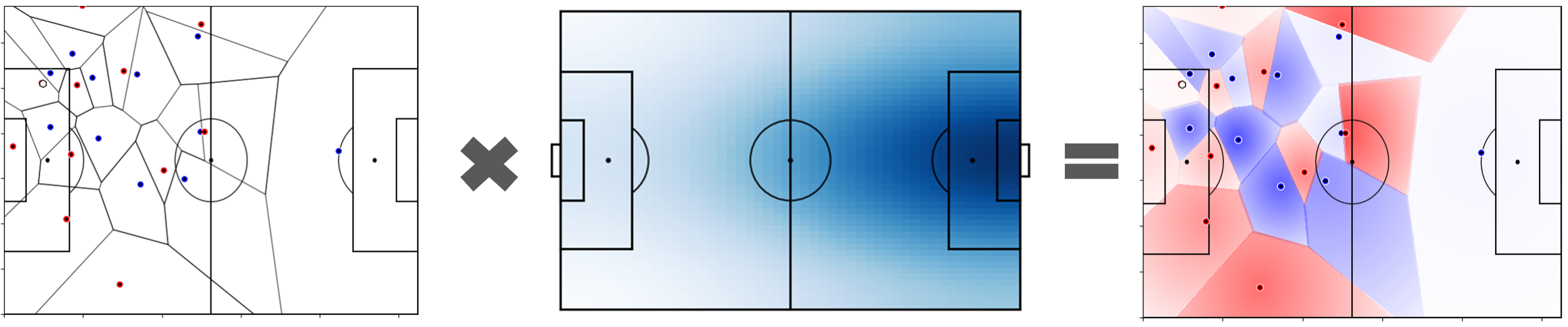}
    \caption{This figure illustrates the procedure for calculating the space score. From left to right, it shows: the Voronoi diagram considering each player's position and velocity; the importance of each area on the pitch; and the final space score distribution, which is the product of the two. The area importance is modeled based on proximity to the opponent's goal and distance from the center of the pitch, and it is maximized in the central area in front of the opponent's goal}
    \label{fig:space_score}
\end{figure}

Fifth, EDMS includes the change in space score when the player moves one meter in each of eight directions. 
Let $\mathcal{D}$ denote the set of eight movement directions:
\begin{equation}
    \mathcal{D}
    =
    \{
    \mathrm{Right},
    \mathrm{Up\text{-}Right},
    \mathrm{Up},
    \mathrm{Up\text{-}Left},
    \mathrm{Left},
    \mathrm{Down\text{-}Left},
    \mathrm{Down},
    \mathrm{Down\text{-}Right}
    \}.
\end{equation}
For each direction $d \in \mathcal{D}$, the change in space score is defined as
\begin{equation}
    \Delta \mathrm{SpaceScore}_{i,t}^{d}
    =
    \mathrm{SpaceScore}(p_{i,t} + u_d)
    -
    \mathrm{SpaceScore}(p_{i,t}),
\end{equation}
where $u_d$ is a one-meter displacement vector in direction $d$.

Finally, EDMS defines a pass score that summarizes the quality of a potential pass to the off-ball player. 
The pass score is computed as a weighted sum of distance, space value, pressure on the receiver, and pressure on the passing lane:
\begin{equation}
    \mathrm{PassScore}_{i,t}
    =
    k_1 d^{\mathrm{ball}}_{i,t}
    +
    k_2 \mathrm{SpaceScore}_{i,t}
    +
    k_3 \tau^{\mathrm{player}}_{i,t}
    +
    k_4 \tau^{\mathrm{passline}}_{i,t}.
\end{equation}
In this study, the weights are set as
\begin{equation}
    k_1 = 0.5, \quad
    k_2 = 0.3, \quad
    k_3 = 0.2, \quad
    k_4 = 0.2.
\end{equation}
These weights were determined based on preliminary analysis of pass success prediction.

\subsubsection{On-ball State}

The On-ball State represents the tactical situation of the player in possession of the ball. 
It is divided into Intra-Possession State and Inter-Possession State. 
The Intra-Possession State describes situations in which the same team maintains possession, while the Inter-Possession State describes situations involving a possible transition of possession between teams.

The Intra-Possession State includes variables related to pressure, goal proximity, dribbling opportunities, shooting opportunities, and long-ball options.

First, EDMS computes the time required for the nearest opposing player to reach the ball:
\begin{equation}
    \tau^{\mathrm{ball}}_{t}.
\end{equation}
This variable represents the pressure applied to the ball carrier.

Second, the distance from the ball carrier to the opponent's goal is computed as
\begin{equation}
    d^{\mathrm{goal}}_{t}
    =
    \|p_{b,t} - p^{\mathrm{goal}}\|_2,
\end{equation}
where $p^{\mathrm{goal}}$ denotes the center of the opponent's goal.

Third, the angle from the ball carrier to the opponent's goal is computed. 
This angle captures the geometric quality of the shooting or attacking situation.

Fourth, EDMS defines a dribble score. 
The dribble score is computed as the change in space score when the ball carrier moves one meter in each of the eight directions:
\begin{equation}
    \mathrm{DribbleScore}^{d}_{t}
    =
    \mathrm{SpaceScore}(p_{b,t} + u_d)
    -
    \mathrm{SpaceScore}(p_{b,t}),
    \quad d \in \mathcal{D}.
\end{equation}
This variable evaluates whether a short movement by the ball carrier would improve the tactical situation.

Fifth, EDMS includes a shot score. 
The shot score evaluates the quality of a potential shot by considering the angle to the goal and the probability that defenders block the shooting lane. 
Only defenders located inside the triangular region formed by the ball carrier and the two goalposts are considered, excluding the goalkeeper. 
The blocking probability is aggregated over the goal mouth, and the shot score is computed only when the ball carrier is within 30 meters of the opponent's goal.

Sixth, EDMS includes a long-ball score. 
The pitch is divided into left, central, and right zones, and the zone containing the tallest attacking player is assigned a high value. 
This variable allows the model to represent long-ball options toward aerially dominant players.

\subsubsection{Inter-Possession State}

The Inter-Possession State represents situations in which possession may change between teams. 
This component is used to model transitional situations, such as turnovers, loose balls, and defensive recoveries.

The Inter-Possession State includes the following variables. 
First, EDMS computes the time required for the nearest player from each team to reach the ball. 
This variable captures which team is more likely to gain possession.

Second, EDMS computes the distance from the ball to each team's goal:
\begin{equation}
    d^{\mathrm{goal}}_{c,t}
    =
    \|p_{b,t} - p^{\mathrm{goal}}_{c}\|_2,
    \quad c \in \{\mathrm{attack}, \mathrm{defense}\}.
\end{equation}

Third, EDMS computes the angle from the ball to each team's goal. 
Fourth, the velocity of the ball is included to capture whether the ball is moving toward a dangerous area or away from pressure. 
Fifth, a transition indicator is included. 
This variable is set to $1$ when the player is switching from attack to defense and $0$ when the player is switching from defense to attack.

\subsubsection{Summary of EDMS}

Table~\ref{tab:edms_summary} summarizes the main components of EDMS.

\begin{table}[t]
    \centering
    \caption{Summary of the EDMS components.}
    \label{tab:edms_summary}
    \begin{tabular}{lll}
        \hline
        Category & Subcategory & Main variables \\
        \hline
        Absolute State 
        & -- 
        & Offside-line distance, formation \\

        Relative State 
        & Off-ball State 
        & \begin{tabular}[t]{@{}l@{}}
            Ball distance, pressure, pass-line pressure, \\
            space score, pass score
          \end{tabular} \\

        Relative State 
        & On-ball / Intra-Possession State 
        & \begin{tabular}[t]{@{}l@{}}
            Ball pressure, goal distance, goal angle, \\
            dribble score, shot score, long-ball score
          \end{tabular} \\

        Relative State 
        & On-ball / Inter-Possession State 
        & \begin{tabular}[t]{@{}l@{}}
            Time to reach ball, goal distance, goal angle, \\
            ball velocity, transition indicator
          \end{tabular} \\
        \hline
    \end{tabular}
\end{table}

Compared with PVS, EDMS explicitly encodes tactical concepts that are relevant to football decision-making. 
This allows the reinforcement learning model to evaluate not only the physical location of players but also the meaning of their positions in the context of pressure, space, passing options, and goal threat. 
Moreover, because EDMS is modular, additional variables such as body orientation, player-specific passing ability, score difference, or set-piece-specific features can be incorporated in future extensions.

\end{document}